\documentclass[lettersize,journal]{IEEEtran}
\usepackage{amsmath,amsfonts}
\usepackage{algorithmic}
\usepackage{algorithm}
\usepackage{array}
\usepackage[caption=false,font=normalsize,labelfont=sf,textfont=sf]{subfig}
\usepackage{textcomp}
\usepackage{stfloats}
\usepackage{url}
\usepackage{verbatim}
\usepackage{graphicx}
\usepackage{cite}
\usepackage{xcolor}
\usepackage{subfig}
\usepackage[color=blue,commandnameprefix=always]{changes}
\usepackage{ulem}

\DeclareMathOperator*{\argmin}{\arg\min}

\newcommand{\bluerow}[1]{%
    \def\temp{}%
    \foreach \x in {#1} {%
        \ifx\temp\empty%
            \xdef\temp{\noexpand\textcolor{blue}{\x}}%
        \else%
            \xdef\temp{\temp & \noexpand\textcolor{blue}{\x}}%
        \fi%
    }%
    \temp%
}
\newcolumntype{B}{>{\color{blue}}c}

\begin{document}
%163.180.116.44
\title{A Sample Article Using IEEEtran.cls\\ for IEEE Journals and Transactions}
\title{Exploring Kernel Transformations for Implicit Neural Representations}

\author{Sheng Zheng, Chaoning Zhang*,~\IEEEmembership{Senior Member,~IEEE}, Dongshen Han, Fachrina Dewi Puspitasari, \\ Xinhong Hao, Yang Yang,~\IEEEmembership{Senior Member,~IEEE}, Heng Tao Shen,~\IEEEmembership{Fellow,~IEEE}
        % <-this % stops a space
\thanks{Sheng Zheng and Chaoning Zhang are with the Center for Future Media and the School of Computer Science and Engineering, University of Electronic Science and Technology of China, Chengdu, China (email: zszhx2021@gmail.com; chaoningzhang1990@gmail.com). Xinhong Hao is with the School of Mechatronical Engineering, Beijing Institute of Technology, Beijing, China (email: haoxinhong@bit.edu.cn). 
Dongshen Han and Fachrina Dewi Puspitasari are with the School of Computing, Kyung Hee University, Yongin-si, Korea (email: han-0129@khu.ac.kr; puspitasaridewi@outlook.com). Yang Yang is with the Center for Future Media and the School of Computer Science and Engineering, University of Electronic Science and Technology of China, Chengdu, China, and also with the Institute of Electronic and Information Engineering, University of Electronic Science and Technology of
China, Guangdong, China (e-mail: dlyyang@gmail.com). Heng Tao Shen is with the Center for Future Multimedia and the School of Computer Science and Engineering, University of Electronic Science and Technology of China, Chengdu, China, and also with the Peng Cheng
Laboratory, Shenzhen, China (e-mail: shenhengtao@hotmail.com).

Corresponding Author: Chaoning Zhang.
}% <-this % stops a space
}

\markboth{Journal of \LaTeX\ Class Files,~Vol.~14, No.~8, August~2021}%
{Shell \MakeLowercase{\textit{et al.}}: A Sample Article Using IEEEtran.cls for IEEE Journals}


\maketitle

\begin{abstract}
Implicit neural representations (INRs), which leverage neural networks to represent signals by mapping coordinates to their corresponding attributes, have garnered significant attention. They are extensively utilized for image representation, with pixel coordinates as input and pixel values as output. In contrast to prior works focusing on investigating the effect of the model's inside components (activation function, for instance), this work pioneers the exploration of the effect of kernel transformation of input/output while keeping the model itself unchanged. A byproduct of our findings is a simple yet effective method that combines scale and shift to significantly boost INR with negligible computation overhead. Moreover, we present two perspectives, depth and normalization, to interpret the performance benefits caused by scale and shift transformation. Overall, our work provides a new avenue for future works to understand and improve INR through the lens of kernel transformation.

\end{abstract}


\section{Introduction}

\IEEEPARstart{I}{mplicit} neural representations (INRs) ~\cite{sitzmann2020implicit, tancik2020fourier}, offering a new paradigm for signal representation using neural networks, have attracted significant attention in the past few years. Unlike the traditional methods that discretize inputs into points cloud, voxel grids, or meshes~\cite{guo2020deep, dai2019adaptive, sun2015hems, ogawa2011missing, wang2022survey, zang2023lce, an2023sp}, INRs define signals in an implicitly defined, continuous, and differentiable manner, making a substantial evolution in this field. This approach, which involves continuously representing signals, aligns with the fundamental principle that real-world signals are inherently continuous. INRs are widely used for image representation by leveraging multi-layer perceptions (MLPs) to fit and model the images~\cite{liu2023finer, xie2023diner, saragadam2022miner, shen2022nerp, dupont2021coin, guo2024compression, rivas2023ice, su2022inras, xu2023implicit, michalkiewicz2019implicit}. The paradigm of INRs to represent images has been demonstrated to provide multiple advantages~\cite{molaei2023implicit, hao2022implicit, chen2019learning, park2019deepsdf, liu2020dist}. Due to their continuous nature, INRs are capable of representing images in a resolution-agnostic manner, enabling them to obtain values at coordinates that lie between pixels. In addition, the required storage memory is not limited by the image resolution, as a result of which, image INRs are both effective and efficient in terms of memory usage~\cite{guo2020deep,wang2022survey}.

When it comes to image representation, INRs still struggle to model the fine details~\cite{sitzmann2020implicit, genova2019deep}. To address this issue, prior studies primarily focus on two approaches: incorporating additional structure within the model~\cite{mildenhall2021nerf, aftab2022multi} and modifying the activation function~\cite{sitzmann2020implicit, genova2019deep, jiang2020local, ramasinghe2022beyond}. Regarding incorporating additional structure within the model, both theoretical derivation and practical experiments suggest that incorporating positional embeddings into the INR model helps in capturing fine details~\cite{mildenhall2021nerf}. Similar ideas have been proposed in MLP-RBF~\cite{sitzmann2020implicit}, while MLP-RBF adopts radial basis functions to encode the positional embedding. Additionally, incorporating multiple heads into the body structure of the INR model is also used to enhance detail capture~\cite{aftab2022multi}. The main body structure captures the overall global features of the image signal, while the additional heads focus on reconstructing separate local details. On the other hand, modifying the activation function used by SIREN involves using the periodic sine activation functions instead of ReLU, which is one of the seminal works of INRs. Similar to SIREN, another study~\cite{ramasinghe2022beyond} utilizes the Gaussian activation function in place of ReLU to investigate its effect. Building on the impressive performance of sinusoidal and Gaussian functions in INRs, Wire~\cite{saragadam2023wire} leverages the strengths of both by incorporating Gabor wavelets. 
A literature review of INR shows that prior works mainly explore how the model's inside components affect its performance.

In essence, image INRs are designed to map coordinates (Input) to their corresponding attributes (Output). Given the I/O mapping nature of this task, we explore the potential of using kernel transformations on both the model's input and output. Instead of focusing on adjusting the model's internal components, such as the activation function, our work thoroughly investigates the impact of kernel transformations applied to the model's input and output. We find that nonlinear kernels harm performance while linear transformations can be beneficial. The effect of linear kernels, including scale and shift transformation, is further explored. Regarding this, our main findings are summarized as follows: the model input benefits from the scale transformation when the scale factor is set to larger than the default 1; the model output gains from the scale transformation when the shift factor is adaptively set to the average value of the target image. We conjecture that straightforwardly combining the two linear transformations, \underline{s}cale-and-\underline{s}hift (SS), can further enhance performance, which is confirmed on multiple INR backbones. Moreover, we evaluated the effectiveness of SS-INR in various experimental setups.  Overall, our contributions are summarized as follows:

\begin{itemize}

\item Recognizing that previous works mainly focus on investigating the model’s inside components, we shift the attention to the model's input and output by utilizing kernel transformations.

\item With a comprehensive study on kernel transformations, we identify scale transformation
for the model's input and shift transformation for the model's output, despite their simplicity, can enhance the performance.

\item As a byproduct of our investigation findings, a new INR framework (termed SS-INR) is proposed, showing competitive performance in diverse setups.

\end{itemize}

As a pioneering attempt to study the effect of kernel transformation on INR performance, this work focuses on empirical findings instead of justified motivation. Nonetheless, we provide two perspectives, model depth and data normalization, to interpret why the scale-and-shift transformation might boost the INR performance. This is expected to inspire future works to continue our investigation.

\section{Related Work}
\label{sec:related_work}

\subsection{Implicit Neural Representations} 

INRs~\cite{sitzmann2020implicit, tancik2020fourier} have emerged as an active research, with applications spanning a wide range of tasks~\cite{liu2024seif, zheng2021pamir, chabra2020deep, mescheder2019occupancy, wu2021irem, yang2021towards}, such as images~\cite{saragadam2023wire, liu2023finer}, and audio signals~\cite{szatkowski2023hypernetworks}. 
Unlike the traditional image recognition methods~\cite{xie2020feature, zhang2020reliability, sun2020fast, wu2025class, zhang2024adversarial}, which use explicit neural networks with images as input, INRs employ implicit neural networks to map the pixel coordinates to pixel values. SIREN~\cite{sitzmann2020implicit}, a notable breakthrough in this domain, leverages a periodic activation sine function instead of the ReLU function to significantly improve the performance of INR. 
The INR models typically face a practical challenge known as spectral bias, which makes it challenging to capture the details of the fine details in signals in comparison to traditional discrete approaches~\cite{leshno1993multilayer, liu2019point, schwarz2022voxgraf, deng2023compressing, kato2018neural, pavllo2020convolutional, luo2021diffusion}.
To alleviate this issue, numerous studies~\cite{tancik2020fourier, xie2023diner, aftab2022multi, genova2019deep, jiang2020local, sitzmann2020implicit, saragadam2023wire, mildenhall2021nerf} have improved the performance of INRs by modifying model structure, such as incorporating positional embeddings and selecting more effective activation functions. Unlike these prior works that concentrate on modifying the model's internal structure, our work explores the model's input and output by kernel transformation.

\subsection{Kernel Transformation Methods} 

The kernel transformation method~\cite{hofmann2008kernel} has been incredibly successful in machine learning, serving as a powerful tool for uncovering nonlinear patterns in data. As deep learning has evolved, numerous works~\cite{huang2006large, niu2012novel, chagas2020classification, wolpert1992stacked, damianou2013deep} have investigated kernel methods in deep models. One family of research~\cite{huang2006large, niu2012novel, chagas2020classification, tang2013deep, zareapoor2018kernelized} focuses on combining deep modules with kernel machines such as SVM, using deep modules as the front end and kernel machines as the back end. Another family of research~\cite{wolpert1992stacked, damianou2013deep, wang2017deep} aims to incorporate kernel methods into deep-stacked architectures, which are designed based on the stack generalization principle~\cite{wolpert1992stacked}. In our work, we delve into using the kernel transformation method for the inputs and outputs of INR.

\section{Preliminary}

\textbf{Image INRs.} The task of image INR is to learn a function $f(\cdot, \theta): \mathbb{R}^{D_{m}} \rightarrow \mathbb{R}^{D_{n}}$, which maps pixel coordinates to pixel values with multi-layer perceptrons (MLPs). Given an input $\mathbf{x}$ and a set of learnable parameters $\theta$,  
the goal of this function is to approximate the ground truth image $\mathbf{Y}$ with $\mathbf{y}$. With an $N$-layer MLP to model $f(\cdot, \theta)$, the output of $n$-{th} layer of MLPs is defined as follows:

\begin{equation}
 \mathbf{y}_{n} = \sigma (\mathbf{W}_{n}\mathbf{y}_{n-1} + \mathbf{b}_{n}), 
\label{eq:mlp}
\end{equation}
where $\sigma$ is the activation function; $\mathbf{W}_{n}$ and $\mathbf{b}_{n}$ are the parameters of $n$-th layer, representing weight and bias, respectively. For $n = 0$, $\mathbf{y}_{0} = \sigma (\mathbf{W}_{0}\mathbf{x} + \mathbf{b}_{0})$ with the pixel coordinates $\mathbf{x}$ as input. The model parameter $\theta$ optimization is formulated as follows:

\begin{equation}
    \theta^* = \argmin_{\theta}\mathcal{L}(f(x, \theta), Y), 
\label{eq:loss}
\end{equation}
where loss function $\mathcal{L}$ is  mean square error (MSE).

\textbf{Evaluation metrics.} In INR tasks, it is common to use the peak-signal-to-noise ratio (PSNR) and structural similarity index measure (SSIM)~\cite{wang2004image} as the quantitative metrics. Thus, we adopt these two metrics to evaluate our proposed method. PSNR mainly measures how faithful the generated images match the original ones, while the SSIM evaluates their structural similarity based on the fundamental principle that humans are sensitive to structural information.

\section{Kernel Transformation for INR}  
\label{sec:method}

\begin{figure*}[!t]
     \centering
     \includegraphics[width=0.75\textwidth]{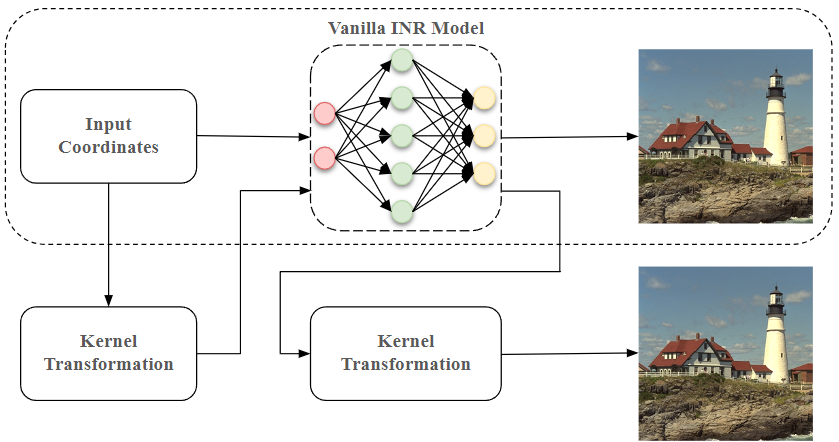}
    \caption{Overview of kernel transformation in INR for image representation. Prior works primarily focus on the effect of the model's internal components, while our work shifts the attention to the model's Input/Output by applying kernel transformations.
    }    
    \label{fig:framework}
\end{figure*}

As discussed in Section~\ref{sec:related_work}, prior works~\cite{mildenhall2021nerf,sitzmann2020implicit, tancik2020fourier,yuce2022structured} on INR for image representation mainly focus on improving the MLP model architecture. For example, the seminal work SIREN~\cite{sitzmann2020implicit} proposes to modify its activation function by replacing the ReLU with periodic sine functions, which significantly facilitates the model to capture high-frequency details. In essence, INR in image representation maps the image coordinates (Input) to their corresponding pixel values (Output). Considering the I/O mapping nature of this task, we delve into whether the transformation of input and output helps the image INR in the mapping function. The overview of kernel transformation in INR is shown in Figure~\ref{fig:framework}. We propose a framework that consists of the default MLP model and I/O transformation modules. Existing INR frameworks, like SIREN, can be seen as a special case of our proposed framework by setting the I/O transformation modules to identity mapping. Therefore, our investigation is complementary to existing works that aim to improve the model architecture.

\textbf{Kernel transformation.} Kernel transformation method, one of the transformation techniques, constructs a mapping function $\phi$ from the input space to the feature space, specifically a reproducing kernel Hilbert space (RKHS)~\cite{hofmann2008kernel}. It has been widely used in various machine-learning problems, where the kernel function determines this mapping, and even a relatively simple linear kernel can significantly enhance the performance. In this work, we utilize kernel methods to transform our input data. Since our task involves image representation, we simply apply the kernel function to map the input. Given the kernel function that models the mapping $f_{m}(z)$, where $z$ is the input of this mapping, we utilize coordinates of INR as the input. It is widely known that the effectiveness of kernel methods largely hinges on the choice of an appropriate kernel function and, consequently, finding an appropriate feature space. Despite the importance of kernel methods, it is still difficult to find a mechanism to guide kernel learning and selection. To this end, we investigate different kernel methods, including nonlinear kernels and linear transformations. To facilitate the investigation, we consistently use the Kodim dataset~\cite{ali2013kodak} that comprises 24 images.

\subsection{Nonlinear Kernels}

First, we investigate the effect of different nonlinear kernels on the input and output. Specifically, we explore five types of nonlinear kernel methods: polynomial kernel, Gaussian kernel, radial kernel, exponential kernel, and Laplacian kernel, which are typically used in machine learning~\cite{hofmann2008kernel}. For \textbf{polynomial kernel}, it can be formulated as follows:
\begin{equation}
k(x) = (a\mathbf{x} + c)^{d}, 
\label{eq:polynomial}
\end{equation}
where $a$, $c$, and $d$ are the hyperparameters. The $d$ is larger than 1, and in our work, we set the $d$ to 3. Gaussian, exponential, and Laplacian kernels all have an exponential form but differ in their underlying distance metrics and smooth properties. For \textbf{Gaussian kernel}, it is defined as:
\begin{equation}
k(x) = exp({-\frac{||\mathbf{x}||^{2}}{2\sigma^{2}}}).
\label{eq:gaussian}
\end{equation}
For \textbf{radial kernel}, it is defined as:
\begin{equation}
k(x) = exp(-\gamma||\mathbf{x}||^{2}). 
\label{eq:radial}
\end{equation}
For \textbf{exponential kernel}, it is defined as:
\begin{equation}
k(x) = exp({-\frac{||\mathbf{x}||}{2\sigma^{2}}}). 
\label{eq:exponential}
\end{equation}
For \textbf{Laplacian kernel}, it is defined as: 
\begin{equation}
k(x) = exp({-\frac{||\mathbf{x}||}{\sigma}}). 
\label{eq:laplacian}
\end{equation}
The hyperparameter $\sigma$ is set to the same value for the Gaussian, exponential, and Laplacian kernels.

We experiment with the above five nonlinear kernels by applying them to the model's input and output, with results shown in Table~\ref{tab:scale_shift_nonlinear}. We observe that these nonlinear kernel methods consistently lead to a decline in performance, regardless of whether it is applied to the input or output. It suggests that applying these nonlinear kernels to the model's input and output has a negative impact on the performance of INRs.

\begin{table}[!htbp]
    \caption{Comparison of different nonlinear kernel methods on the INR performance when applied separately to the model's input and output.}
    \label{tab:scale_shift_nonlinear}
    \centering
    \resizebox{0.45\textwidth}{!}{%
    \begin{tabular}{ccccccccc}
        \hline
         & \multicolumn{2}{c}{Input} & \multicolumn{2}{c}{Output} \\
        Methods &  PSNR &  SSIM  & PSNR & SSIM \\
        \hline
        Vanilla  &35.40 & 0.9535 & 35.40 & 0.9535  \\
        Polynomial &24.48 &0.7209 & 33.46 & 0.9194 \\
        Laplacian  & 15.38 & 0.3966& 14.36 & 0.4650  \\
        Gaussian  &15.39 &0.3967 & 14.41& 0.4712\\
        Exponential  & 15.35 & 0.3960& 14.17 & 0.4582  \\
        Radial & 15.38 & 0.3966 & 12.40 & 0.3891 \\
        \hline
    \end{tabular}%
    }
\end{table}

\subsection{Linear Transformations}
\label{subsec:linear_kernel_transformation}

Given that the above investigated nonlinear kernels consistently decrease performance, we shift the focus from nonlinear kernels to linear transformation. Specifically, we explore the effect of scale transformation and shift transformation.

\begin{table}[!t]
    \caption{Effect of scale factors on the INR performance when a scale transformation is applied to the model's output.}
    \label{tab:output_scale}
    \centering
    \resizebox{0.45\textwidth}{!}{%
    \begin{tabular}{cccccccccc}
        \hline
        Scale factor& 0.25 & 0.5 & 1 & 2 & 4 \\
        \hline
        PSNR &31.03 &32.92 & 33.58 & 33.24 & 31.29 \\
        SSIM &0.8938 &0.9156 & 0.9202 & 0.9161 &0.8833 \\
        \hline
    \end{tabular}%
    }
\end{table}

\begin{figure}[!t]
\centering
{\includegraphics[width=0.90\linewidth]{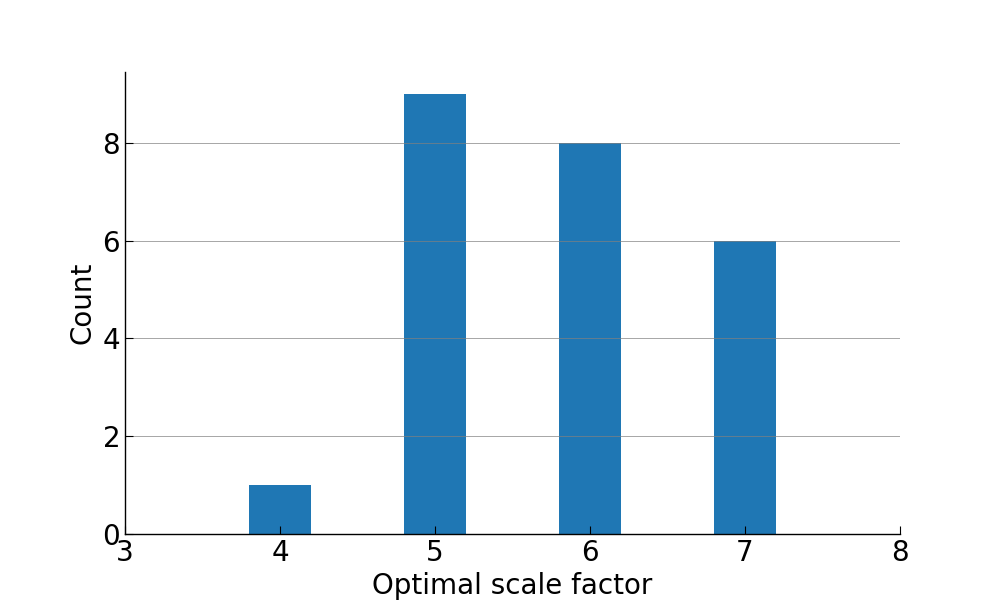}}~
\caption{Count of optimal scale factors for 24 studied images.}
\label{fig:scale_factor_input_distribution}
\end{figure}

\begin{figure}[!t]
\centering
{\includegraphics[width=0.90\linewidth]{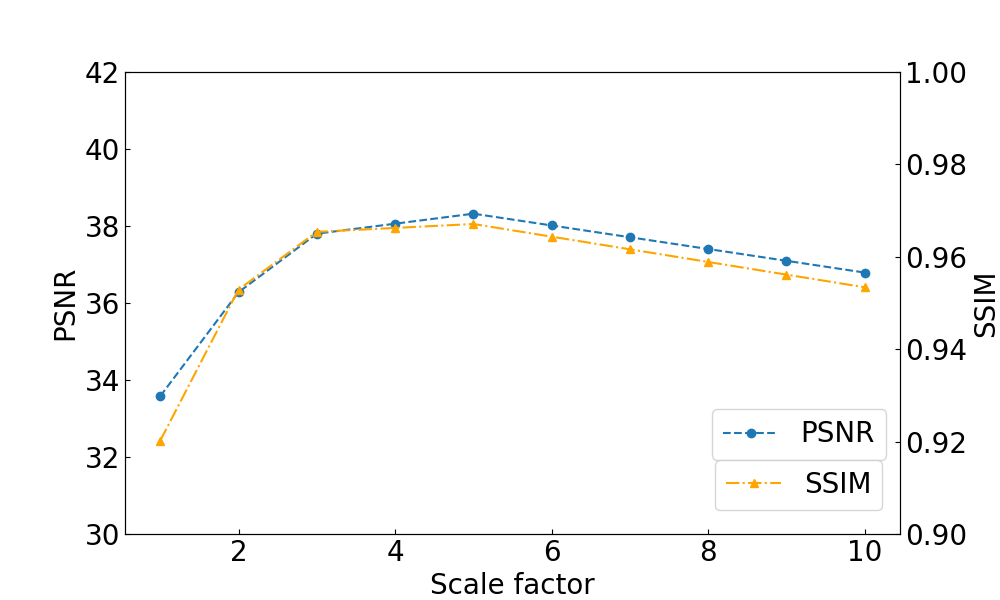}}~
\caption{Effect of scale factors on the INR performance when a scale transformation is applied to the model's input.}
\label{fig:scale_factor_input}
\end{figure}

\textbf{Scale transformation.} We experiment with different scale factors for the input and output kernel transformation. 
As shown in Table~\ref{tab:output_scale}, the performance of INR declines when the scale factor of output kernel transformation is adjusted away from the default 1, whether it is increased or decreased. This trend aligns with the performance observed for each image. We further explore how the scale factor affects the input kernel transformation, and the results in Figure~\ref{fig:scale_factor_input_distribution} show that it varies across different images, with scale factors larger than 1 being beneficial. 
The optimal scale factors typically fall between 4 and 7, with the majority clustering around 5 and 6. Specifically, out of a total of 24 samples, nine samples perform best with a scaling factor of 5, and 8 samples achieve their best results with a scaling factor of 6.
We then apply a scale factor ranging from 1 to 10 to the Kodim dataset images to find the optimal scale factor for the whole dataset, with the results shown in Figure~\ref{fig:scale_factor_input}.  
We observe that there is a noticeable boost in performance when the scale factor is larger than 1, with a scale factor value of 5 performing best. Therefore, we set the scale factor to 5.

\begin{table}[!t]
    \caption{Effect of shift factors on the INR performance when a shift transformation is applied to the model's input.}
    \label{tab:coordinate_shift}
    \centering
    \resizebox{0.45\textwidth}{!}{%
    \begin{tabular}{cccccccccc}
        \hline
        Shift factor & -100 & -10  & 0 & 10 & 100  \\
        \hline
        PSNR & 29.86 &33.38 & 33.58& 33.25& 30.12  \\
        SSIM & 0.8674 & 0.9186 & 0.9202 &0.91668 &0.8709    \\
        \hline
    \end{tabular}%
    }
\end{table}

\begin{table}[!t]
    \caption{Effect of shift factors on the INR performance when a shift transformation is applied to the model's output.}
    \label{tab:output_shift}
    \centering
    \resizebox{0.45\textwidth}{!}{%
    \begin{tabular}{cccccccccc}
        \hline
        Shift factor&  -0.5 &-0.2 & 0 & 0.2 & 0.5 \\
        \hline
        PSNR& 32.72& 34.02& 33.58 &30.54 &30.03 \\
        SSIM& 0.9109& 0.9268& 0.9202 &0.8996 &0.8651  \\
        \hline
    \end{tabular}%
    }
\end{table}

\begin{figure}[!t]
\centering
{\includegraphics[width=0.90\linewidth]{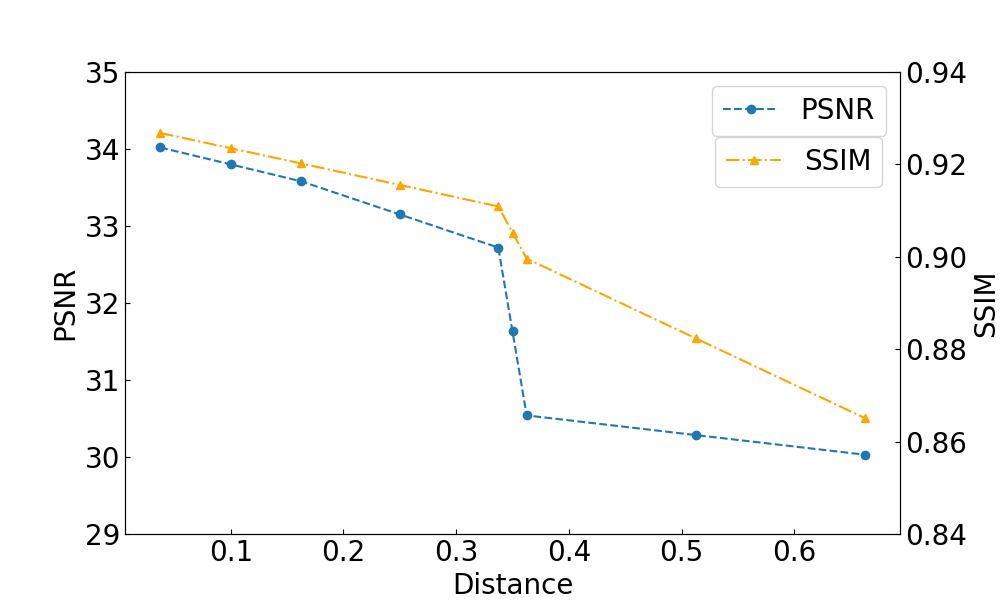}}~
\caption{The relationship between the distance and corresponding PSNR/SSIM for shift transformation on the model's output. This distance is defined as the difference between the applied shift value and the average value of all target images in the dataset.}
\label{fig:bias}
\end{figure}

\begin{table}[!t]
    \caption{Effect of adaptive shift factor on the INR performance when a shift transformation is applied to the model's output.}
    \label{tab:output_adaptive}
    \centering
    \resizebox{0.35\textwidth}{!}{%
    \begin{tabular}{cccccccccc}
        \hline
        Methods & PSNR &  SSIM  \\
        \hline
        Baseline & 33.58 & 0.9202 \\
        Shift factor (w/o adaptive) & 34.02 & 0.9268  \\
        Shift factor (w/ adaptive) & 35.16 & 0.9402  \\
        \hline
    \end{tabular}%
    }
\end{table}

\textbf{Shift transformation.} In addition to examining the impact of the scale transformation, we also investigate how the shift transformation affects the input and output kernel transformations. As shown in Table~\ref{tab:coordinate_shift}, applying the shift factor to the input kernel transformation tends to reduce performance.
When it comes to shift transformation on output, a shift factor of -0.2 improves performance (see Table~\ref{tab:output_shift}), which inspires our further investigation. Specifically, we apply a grid search of the optimal shift value for all the investigated images. Empirically, we find that there exists a negative correlation between the PSNR and the distance between the applied shift value and the average value of whole target images (see Figure~\ref{fig:bias}). Our empirical findings reveal a negative correlation between PSNR and the distance, where this distance is defined as the difference between the applied shift value and the average value of all target images in the dataset, measuring the relationship between the shift value and this average. It suggests that the optimal shift factor is related to the average value of whole target images. Since the INR task aims to fit each image individually, we propose to dynamically adjust this shift factor based on the average value of each specific image. In other words, we adaptively use the average value of each target image as the shift factor. As shown in Table~\ref{tab:output_adaptive}, using this adaptive shift factor for output kernel transformation improves performance. Overall, applying a shift factor, set adaptively to the average pixel value of the target image, on the model's output can significantly enhance the INR performance.
\begin{figure*}[!t]
    \centering
    \captionsetup[subfigure]{justification=centering}
    % \subfloat[Ground truth]
    {\includegraphics[width=0.65\textwidth]{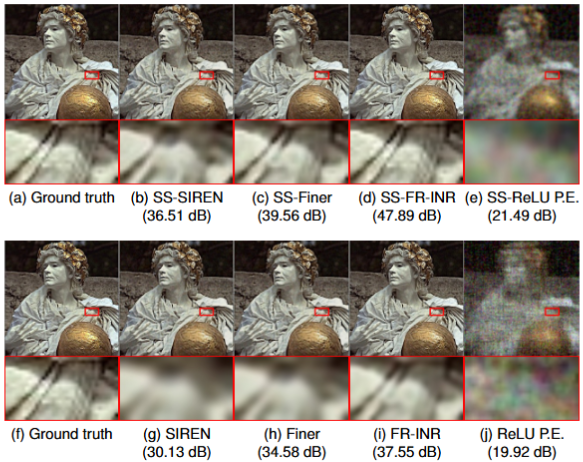}}%
    % \hfil
    % \subfloat[SS-SIREN \\ (36.51 dB)]{\includegraphics[width=0.15\textwidth]{figs_unbias/image_representation/low_resolution/siren_scale_processed.jpg}}%
    % % \hfil
    % \subfloat[SS-Finer \\ (39.56 dB)]{\includegraphics[width=0.15\textwidth]{figs_unbias/image_representation/low_resolution/finer_scale_processed.jpg}}%
    % % \hfil
    % \subfloat[SS-FR-INR \\ (47.89 dB)]{\includegraphics[width=0.15\textwidth]{figs_unbias/image_representation/low_resolution/fr_scale_processed.jpg}}%
    % \subfloat[SS-ReLU P.E. \\ (21.49 dB)]
    % {\includegraphics[width=0.15\textwidth]{figs_unbias/image_representation/low_resolution/nerf_scale_processed.jpg}}%

    % \subfloat[Ground truth ]{\includegraphics[width=0.15\textwidth]{figs_unbias/image_representation/low_resolution/gt_processed.jpg}}%
    % % \hfil
    % \subfloat[SIREN \\ (30.13 dB)]{\includegraphics[width=0.15\textwidth]{figs_unbias/image_representation/low_resolution/siren_processed.jpg}}%
    % % \hfil
    % \subfloat[Finer \\ (34.58 dB)]
    % {\includegraphics[width=0.15\textwidth]{figs_unbias/image_representation/low_resolution/finer_processed.jpg}}%
    % % \hfil
    % \subfloat[FR-INR \\ (37.55 dB)]{\includegraphics[width=0.15\textwidth]{figs_unbias/image_representation/low_resolution/fr_processed.jpg}}%
    % % \hfil
    % \subfloat[ReLU P.E. \\ (19.92 dB)]{\includegraphics[width=0.15\textwidth]{figs_unbias/image_representation/low_resolution/nerf_processed.jpg}}%
    % \hfil
    \caption{Qualitative comparison of vanilla INR backbones with and without SS modules on image fitting. The clothing on the sculpture within the red rectangle provides a more apparent comparison of the clarity of the visual representation of each method. Compared to the vanilla INR backbones, those with SS modules are better at capturing fine details.}
    \label{fig:image_representation}
\end{figure*}

\subsection{Scale-and-Shift INR}

Our above findings indicate that the INR performance benefits from (1) scaling the model's input and (2) shifting the model's output. It is conjectured that combining (1) and (2) can further enhance the performance. We term the proposed new INR method as 
\underline{s}cale-and-\underline{s}hift \underline{i}mplicit \underline{n}eural \underline{r}epresentation (SS-INR). SS-INR consists of a vanilla INR backbone and the SS modules. To confirm our hypothesis, we conduct the experiments of SS-INR on the Kodim dataset~\cite{ali2013kodak}, with the results presented in Table~\ref{tab:scale_and_shift}.  Additionally, we perform experiments on another set of 20 randomly selected images from the DIV2K dataset~\cite{agustsson2017ntire} (see Table~\ref{tab:scale_and_shift}). These results consistently show that (1) and (2) complement each other. Based on multiple types of vanilla INR backbones (ReLU P.E. SIREN, Finer, and FR-INR), the effectiveness of SS-INR is evaluated in diverse setups, with the results detailed in the following section.

\begin{table}[!htbp]
    \caption{Effect of combining scale transformation for model's input and shift transformation for model's output on INR performance using Kidom and DIV2K dataset.}
    \label{tab:scale_and_shift}
    \centering
    \resizebox{0.49\textwidth}{!}{%
    \begin{tabular}{cccccccccc}
        \hline
         & \multicolumn{2}{c}{Kidom}  & \multicolumn{2}{c}{DIV2K} \\
        Methods & PSNR &  SSIM  & PSNR & SSIM \\
        \hline
        Baseline & 33.58 & 0.9202 & 32.37 & 0.9294  \\
        Scale transformation (input) &38.32 & 0.9671 & 37.28 & 0.9692  \\
        Shift transformation (output) & 35.16 & 0.9402 & 34.55 & 0.9562\\
        Scale-and-Shift (input/output)&39.85 & 0.9758 & 39.14 & 0.9758\\
        \hline
    \end{tabular}%
    }
\end{table}

\begin{table*}[!t]
    \caption{Quantitative comparison of various methods on image fitting.}
    \label{tab:different_methods}
    \centering
    \resizebox{0.90\textwidth}{!}{%
    \begin{tabular}{ccccccccccccc}
        \hline
         Methods & ReLU P.E. & SIREN & Finer  & FR-INR & SS-ReLU P.E.  & SS-SIREN  & SS-Finer  & SS-FR-INR\\
         \hline
         PSNR & 20.75  & 33.58 & 37.16&  39.16 &23.17 & 39.85 & 39.99 & 47.11 \\
         SSIM & 0.3071 & 0.9202 &0.9598&  0.9723 &0.5000 & 0.9758 & 0.9766  & 0.9942 \\
        \hline
    \end{tabular}%
    }
\end{table*}

\begin{figure*}[!t]
    \centering
    \captionsetup[subfigure]{justification=centering}
    % \subfloat[Ground truth]
    {\includegraphics[width=0.65\textwidth]{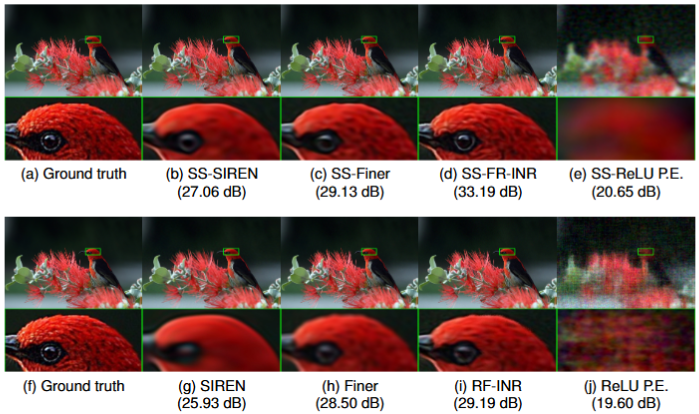}}
    % \subfloat[SS-SIREN \\ (27.06 dB)]{\includegraphics[width=0.18\textwidth]{figs_unbias/image_representation/high_resolution/siren_scale_processed.jpg}}%
    % % \hfil
    % \subfloat[SS-Finer \\ (29.13 dB)]{\includegraphics[width=0.18\textwidth]{figs_unbias/image_representation/high_resolution/finer_scale_processed.jpg}}%
    % % \hfil
    % \subfloat[SS-FR-INR \\ (33.19 dB)]{\includegraphics[width=0.18\textwidth]{figs_unbias/image_representation/high_resolution/fr_scale_processed.jpg}}%
    % % \hfil
    % \subfloat[SS-ReLU P.E. \\ (20.65 dB)]
    % {\includegraphics[width=0.18\textwidth]{figs_unbias/image_representation/high_resolution/nerf_scale_processed.jpg}}%
    % % \hfil

    % \subfloat[Ground truth]{\includegraphics[width=0.18\textwidth]{figs_unbias/image_representation/high_resolution/gt_processed.jpg}}
    % \subfloat[SIREN \\ (25.93 dB)]{\includegraphics[width=0.18\textwidth]{figs_unbias/image_representation/high_resolution/siren_processed.jpg}}%
    % % \hfil
    % \subfloat[Finer \\ (28.50 dB)]
    % {\includegraphics[width=0.18\textwidth]{figs_unbias/image_representation/high_resolution/finer_processed.jpg}}%
    % % \hfil
    % \subfloat[RF-INR \\ (29.19 dB)]{\includegraphics[width=0.18\textwidth]{figs_unbias/image_representation/high_resolution/fr_processed.jpg}}%
    % % \hfil
    % \subfloat[ReLU P.E. \\ (19.60 dB)]{\includegraphics[width=0.18\textwidth]{figs_unbias/image_representation/high_resolution/nerf_processed.jpg}}%
    % \hfil
    \caption{Qualitative comparisons of vanilla INR backbones with and without SS modules on megapixel image fitting. The head of the bird within the red rectangle provides a more apparent comparison of the clarity of the visual representation of each method. Similar to image fitting, vanilla INR backbones with SS modules can capture fine details compared to those without SS modules.
    }
\label{fig:image_representation_high_resolution}
\end{figure*}

\section{Experiments}
\label{sec:experiments}

In this part, we conduct various experiments to verify the effectiveness of our proposed SS modules on different vanilla INR backbones. We also evaluate its performance across various image signals, including natural and CT images. Even though our primary focus is on image representation, we also extended our experiments to explore audio fitting.

\subsection{Image Fitting}

We conduct experiments on four vanilla backbones: ReLU positional encoding (ReLU P.E.)~\cite{mildenhall2021nerf}, SIREN~\cite{sitzmann2020implicit}, Finer~\cite{liu2023finer}, and FR-INR~\cite{shi2024improved}, each enhanced with our proposed SS modules. For simplicity, we refer to SIREN with SS modules as SS-SIREN and similarly for the other backbones. The optimal scale factor varies across different vanilla backbones: 0.3 for ReLU P.E., 5 for SIREN, and 2 for both Finer and FR-INR. To ensure a fair comparison, we configure all INR models with three hidden layers, each containing 256 neurons. These models utilize the same Adam optimizer and mean square error (MSE) loss function. All parameters of these models are set according to the specifications provided in the officially released code. The experiments are conducted on the Kodim dataset~\cite{ali2013kodak}, as mentioned in Section~\ref{sec:method}, with images at a resolution of 256 $\times$ 256. Without loss of generality, we set the training epochs to 500. Table~\ref{tab:different_methods} shows the results for three different vanilla backbones, both with and without the SS modules. The backbones equipped with SS modules perform better, highlighting the effectiveness of our proposed SS modules in capturing fine details. Additionally, Figure~\ref{fig:image_representation} shows the qualitative results with the details of different methods. We can notice that five vanilla backbones powered with SS modules capture more fine details in the image, such as the clothing on the sculpture, performing much better than those without these modules.

Moreover, we compare our proposed method on a bird image with a resolution of 2250 $\times$ 1500, where the high-resolution one can be used to verify the ability of the model to encode high-frequency details. Figure~\ref{fig:image_representation_high_resolution}
visualizes the results with the fitting PSNR. Among all these methods, five vanilla backbones powered with SS modules provide more vivid results with fine details, as highlighted by the red square in Figure~\ref{fig:image_representation_high_resolution}. This indicates that our proposed SS modules consistently enhance the performance of different vanilla backbones in terms of megapixel image representation.

\begin{figure*}[!t]
    \centering
    \captionsetup[subfigure]{justification=centering}
    % \subfloat[Ground truth]{\includegraphics[width=0.17\textwidth]{figs_unbias/CT_reconstruction/gt_processed.jpg}}%
    % \hfil
    % \subfloat[SS-SIREN \\ (40.48 dB)]
    {\includegraphics[width=0.65\textwidth]{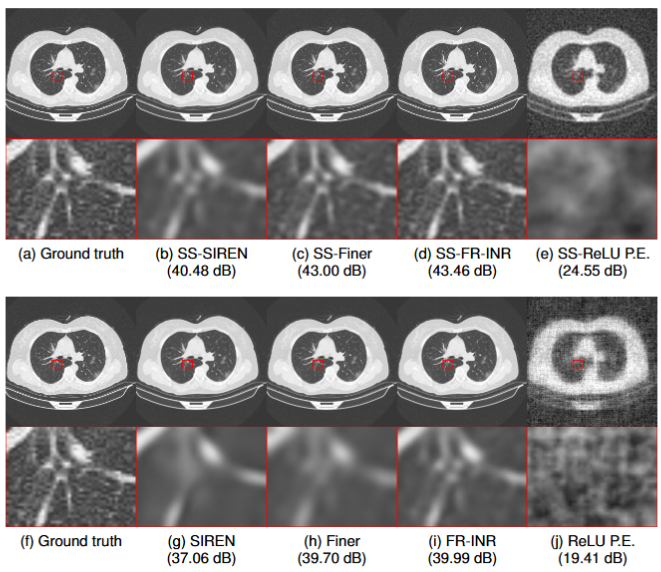}}%
    % \hfil
    % \subfloat[SS-Finer \\ (43.00 dB)]{\includegraphics[width=0.17\textwidth]{figs_unbias/CT_reconstruction/finer_scale_processed.jpg}}%
    % % \hfil
    % \subfloat[SS-FR-INR \\ (43.46 dB)]{\includegraphics[width=0.17\textwidth]{figs_unbias/CT_reconstruction/fr_scale_processed.jpg}}%
    % % \hfil
    % \subfloat[SS-ReLU P.E. \\ (24.55 dB)]{\includegraphics[width=0.17\textwidth]{figs_unbias/CT_reconstruction/nerf_scale_processed.jpg}}%
    % % \hfil

    % \subfloat[Ground truth ]{\includegraphics[width=0.17\textwidth]{figs_unbias/CT_reconstruction/gt_processed.jpg}}%
    % % \hfil
    % \subfloat[SIREN \\ (37.06 dB)]{\includegraphics[width=0.17\textwidth]{figs_unbias/CT_reconstruction/siren_processed.jpg}}%
    % % \hfil
    % \subfloat[Finer \\ (39.70 dB)]{\includegraphics[width=0.17\textwidth]{figs_unbias/CT_reconstruction/finer_processed.jpg}}%
    % % \hfil
    % \subfloat[FR-INR \\ (39.99 dB)]{\includegraphics[width=0.17\textwidth]{figs_unbias/CT_reconstruction/fr_processed.jpg}}%
    % % \hfil
    % \subfloat[ReLU P.E. \\ (19.41 dB)]{\includegraphics[width=0.17\textwidth]{figs_unbias/CT_reconstruction/nerf_processed.jpg}}%
    \caption{Qualitative comparisons of vanilla INR backbones with and without SS modules on CT image reconstruction. The vanilla INR backbones with SS modules demonstrate competitive performance in capturing fine details. 
    }
    \label{fig:ct}
\end{figure*}

\begin{table*}[!t]
    \caption{Quantitative comparisons of various methods on CT image reconstruction.}
    \label{tab:different_methods_ct}
    \centering
    \resizebox{1.0\textwidth}{!}{%
    \begin{tabular}{ccccccccccccc}
        \hline
         Methods & ReLU P.E. & SIREN & Finer & FR-INR & SS-ReLU P.E.  & SS-SIREN  & SS-Finer&  SS-FR-INR \\
         \hline
         PSNR &19.90 &38.20 & 40.82& 41.67 &23.23 & 43.36 & 43.73 &  45.27 \\
         SSIM &0.2685 &0.9399 & 0.9643& 0.9735 & 0.5138& 0.9822 & 0.9828& 0.9871 \\
        \hline
    \end{tabular}%
    }
\end{table*}

\subsection{CT Reconstruction}

We investigate our proposed method for encoding computed tomography (CT) images. Unlike the three-channel color image, CT scans are generally created by capturing information on how X-rays are attenuated as they pass through the body, leading to a grayscale representation of the internal structures of the body. We use the COVIDx-CT dataset~\cite{Gunraj2020} to evaluate the performance of our proposed method, where we randomly choose 20 CT images in this dataset with a resolution of 512 $\times$ 512. We apply different vanilla backbones equipped with SS modules to this task and evaluate their performance. 
The quantitative and qualitative results are presented in Table ~\ref{tab:different_methods_ct} and Figure~\ref{fig:ct}, respectively. These results highlight a significant improvement for these with our SS modules.

\subsection{Audio Representation}

In addition to evaluating our proposed method on image signals, we also conduct experiments on audio signals, which allows us to verify its effectiveness across different signal types. Following~\cite{sitzmann2020implicit}, we utilize the first few seconds of Bach’s Cello Suite No.1 music signal as the audio data. We set the training epochs to 500 and utilize the mean and standard deviation with 10 independent fitting trials as the evaluation metrics to assess our proposed methods. We experiment with ReLU P.E, SIREN, Finer, FR-INR and their corresponding SS-INR, including  SS-ReLU P.E, SS-SIREN, SS-Finer, SS-FR-INR, with the results shown in Table~\ref{tab:audio}. We can observe that our proposed SS modules consistently enhance the performance of different vanilla backbones, suggesting that our proposed method can represent audio signals well.

\begin{table}[!htbp]
    \caption{The quantitative evaluation results of different methods on audio fitting. The mean and variance of the reconstruction MSE are evaluated over ten independent fitting trials.}
    \label{tab:audio}
    \centering
    \resizebox{0.50\textwidth}{!}{%
    \begin{tabular}{ccccccccccccccccccccc}
        \hline
        Methods & MSE Mean  & MSE Standard Dev. \\
        \hline
        ReLU P.E. & $2.621\times10^{-2}$ & $5.258\times10^{-3}$  \\
        SIREN & $1.123\times10^{-3}$ & $8.954\times10^{-5}$   \\
        Finer & $1.010\times10^{-3}$ & $4.111\times10^{-5}$   \\
        FR-INR & $6.201\times10^{-4}$ & $2.185\times10^{-5}$   \\
        \hline
        SS-ReLU P.E. & $2.048\times10^{-2}$ & $5.125\times10^{-3}$  \\
        SS-SIREN& $5.162\times10^{-4}$ & $2.745\times10^{-5}$  \\
        SS-Finer & $5.298\times10^{-4}$ & $1.752\times10^{-5}$  \\
        SS-FR-INR & $2.938\times10^{-4}$ & $1.232\times10^{-5}$   \\
        \hline
        
    \end{tabular}%
}
\end{table}

\section{Discussions}
This work comprehensively explores the influence of the kernel transformation (of the model input and outputs) on the performance of INR. We find that nonlinear kernels harm the model performance, while simple linear transformation can be beneficial. Empirically, we have shown that scale transformation of the input and shift transformation of the output significantly boost the INR performance. It is not the focus of this work to understand why linear transformation affects INR performance. Nonetheless, to inspire future works to continue this exploration, we attempt to provide two perspectives, model depth and data normalization, to interpret the role of scale and shift transformation.

\textbf{Depth perspective.} An astute reader can note that the scale transformation of the input and the shift transformation can be interpreted as additional model layers. In other words, the benefit caused by our proposed linear transformation (LT) might be attributed to the increased model depth. Here, we present the results of modifying the model architecture by adding or removing  
fully connected (FC) and LT for model input and output, shown in Table~\ref{tab:number_hidden_layers}.
Adding more layers to the INR model improves its performance, but it also significantly increases the number of model parameters. However, the INR model with an input and output kernel transformation, i.e., adding LT layers, outperforms the models with adding additional FC layers, with 39.68 dB compared to 37.98 dB. 
Simultaneously, adding an FC layer increases the number of parameters, while adding an LT layer maintains the number of parameters. For instance, an FC layer might expand the total parameters to 0.264 million while an LT layer maintains it at 0.1988 million. This indicates that depth perspective makes it hard to explain the reason that INR performance is enhanced by input and output kernel transformation. 

\begin{table}[!htbp]
    \caption{Effect of removing or adding model layers. FC denotes the fully connected layer, which is the basic layer in an INR model. LT denotes the linear transformation layer proposed in this work. Adding more FC layers leads to higher PSNR but at the cost of a non-trivial increase of model parameters. Adding LT layers leads to yet higher PSNR without increasing the model parameters.}
    \label{tab:number_hidden_layers}
    \centering
    \resizebox{0.50\textwidth}{!}{%
    \begin{tabular}{cccccccccc}
        \hline
        Action (Layers after action) & PSNR &  Parameters  \\
        \hline
        Removing one FC  layer (4) & 31.02 & 0.133M \\
        Baseline (5) & 33.58 & 0.198M  \\
        Adding one FC layer (6) & 34.85& 0.264M  \\
        Adding two FC layers (7)  & 35.92 & 0.330M  \\
        \hline
        Adding two LT layers (7) & 39.85 & 0.198M \\
        \hline
    \end{tabular}%
    }
\end{table}

\begin{table}[!htbp]
    \caption{Effect of normalization for the model's input and output. We present the results across different normalization ranges applied to the models' input and output. The results suggest that the effectiveness of our proposed method can, at least in part, be interpreted from the data normalization perspective.}
    \label{tab:input_output_norm}
    \centering
    \resizebox{0.50\textwidth}{!}{%
    \begin{tabular}{cccccccccc}
        \hline
        Methods & PSNR &  SSIM  \\
        \hline
        Input ($\left[0, 255\right]$)& 29.51 & 0.8409   \\
        Input normalization ($\left[-1, 1\right]$) & 33.58 & 0.9202  \\
        Input normalization ($\left[-5, 5\right]$) & 38.32 & 0.9671  \\
        \hline
        Output ($\left[0, 255\right]$)& 7.62 & 0.1853 \\
        Output normalization ($\left[-1, 1\right]$)& 33.58&  0.9202 \\
        Output normalization (adaptive)& 35.16 & 0.9402\\
        \hline
    \end{tabular}%
    }
\end{table}

\textbf{Normalization perspective.} In the machine learning community, performing normalization on the data is widely known to facilitate the learning process~\cite{ioffe2015batch, ba2016layer}. For example, it is a \textit{de facto} standard practice to normalize the input images to have zero means and unit variances~\cite{lecun2002efficient, wiesler2011convergence}.
Mathematically, such normalization performs the linear transformation of the data. From this perspective, the linear transformation, as a result of our comprehensive search on kernel transformation, can be perceived as the data normalization. The common practice for INR tasks is typically to normalize the image coordinates to the range of $\left[-1, 1\right]$, while it is worth considering whether this is the optimal range. In our work, we find that the performance of INR benefits from the scale transformation with the data normalized to a different range, \textit{i.e.}$, \left[-5, 5\right]$ (see Table~\ref{tab:input_output_norm}). Similarly, our shift transformation for the model outputs (\textit{pixel values}) aligns with the concept of the normalization of image pixel values by removing their mean value. The difference is that the INR task here adaptively removes the mean value of the target image instead of that of all the images in the dataset. Even though the proposed SS-INR method is a byproduct of our study on kernel transformation, its success can, at least partially, be interpreted from the data normalization perspective. We leave a deep study on the reason for its effectiveness to future works.

\section{Conclusion}

Recognizing prior INR works mainly focus on the study of the model's internal components, this work pioneers to investigate transformations on the INR model's I/O. With a comprehensive study on nonlinear kernels and linear transformations, we find that the former harms the INR performance while the latter can give benefit. For the linear kernels, we explore scale and shift transformations and observe that scaling the input with a factor larger than 1 tends to improve performance while shifting the output with adaptive prior benefit. Our investigation naturally results in a new INR method (SS-INR) that combines input scaling and output shifting, which shows consistent performance improvement with four seminal INR backbones on various types of data, including natural images, CT images, and audio signals. Interpreting the linear transformation modules in our proposed SS-INR as additional model layers, we compare our SS-INR with the vanilla with additional layers and find that input scaling and output shifting, despite their simplicity, are more effective than adding more layers. Except for the model depth perspective, we note that the effectiveness of our SS-INR can also be interpreted from the normalization perspective. Since our work primarily focuses on static kernel transformation by offering a thorough empirical study to understand its effectiveness in various tasks, dynamic kernel transformation can be an interesting area for future exploration.

\bibliography{egbib, mixup, reg}
\bibliographystyle{IEEEtran}

% \newpage

\section{Biography Section}

\begin{IEEEbiography}[{\includegraphics[width=1in,height=1.25in,clip,keepaspectratio]{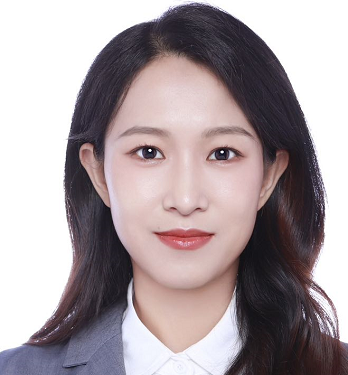}}]{Sheng Zheng} is currently conducting research (intern) at the University of Electronic Science and Technology of China, Chengdu, China, towards obtaining the Ph.D. degree at Beijing Institute of Technology, Beijing, China. Before this, she received a master’s degree in Mechanical Engineering from Dalian University of Technology, Dalian, China, in 2018. Her research interests include computer vision and machine learning.
\end{IEEEbiography}

\begin{IEEEbiography}[{\includegraphics[width=1in,height=1.25in,clip,keepaspectratio]{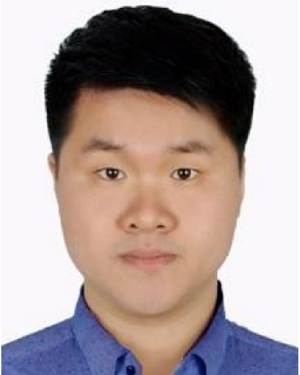}}]{Chaoning Zhang} (Senior Member, IEEE) received the bachelor's and master's degrees in electrical engineering from the Harbin Institute of Technology, Harbin, China, in 2012 and 2015, respectively, and the Ph.D. degree from Korea Advanced Institute of Science Technology (KAIST), South Korea, in 2021. He is currently with the University of Electronic Science and Technology of China, Chengdu, China. His research interests include computer vision, generative AI (AIGC), and emerging topics in AI.
\end{IEEEbiography}

\begin{IEEEbiography}[{\includegraphics[width=1in,height=1.25in,clip,keepaspectratio]{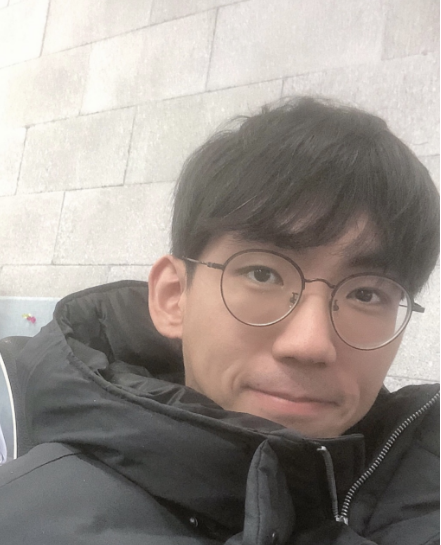}}]{Dongshen Han} received the bachelor's degree in electronic information science and technology from Nanjing Agricultural University, Nanjing, China, in 2018, the master's degree in Electronics Engineering from Nanjing Agricultural University, Nanjing, China, in 2020. He is currently pursuing his Ph.D. degree at Kyung Hee University. His research interests include computer vision and machine learning.
\end{IEEEbiography}

\begin{IEEEbiography}[{\includegraphics[width=1in,height=1.25in,clip,keepaspectratio]{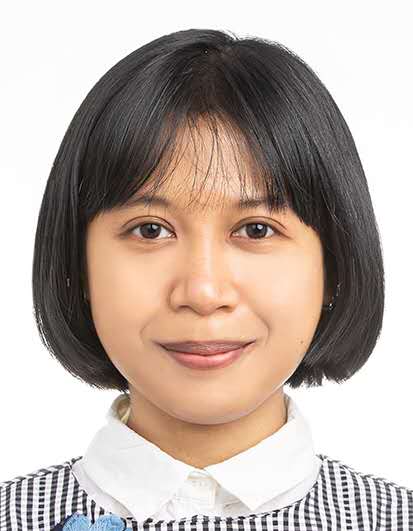}}]{Fachrina Dewi Puspitasari} received a B.Eng. in Mechanical Engineering
from Universitas Indonesia, in 2014 and an MSc. in Industrial and Systems Engineering from KAIST, in 2024. She is currently pursuing his Ph.D. degree at Kyung Hee University. Her research interests mainly include computer vision and large language models.
\end{IEEEbiography}

\begin{IEEEbiography}[{\includegraphics[width=1in,height=1.25in,clip,keepaspectratio]{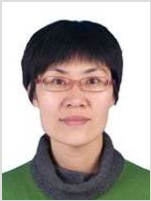}}]{Xinhong Hao} received the Ph.D. degree in the School of Mechatronic Engineering from 
Beijing Institute of Technology, Beijing, China, in
2007. She is currently a professor at
the Beijing Institute of Technology. Her main research interests include computer vision and signal processing.
\end{IEEEbiography}

% % \vspace{30}
\begin{IEEEbiography}[{\includegraphics[width=1in,height=1.25in,clip,keepaspectratio]{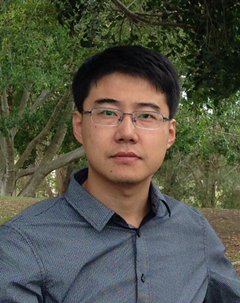}}]{Yang Yang} (Senior Member, IEEE) received the bachelor’s degree in computer science from Jilin University, Changchun, China, in 2006, the master’s degree in computer science from Peking University, Beijing, China, in 2009, and the Ph.D. degree in computer science from The University of Queensland, Brisbane, QLD, Australia, in 2012. He is currently with the University of Electronic Science and Technology of China, Chengdu, China. His research interests include multimedia content analysis, computer vision, and social media analytics.
\end{IEEEbiography}

\begin{IEEEbiography}[{\includegraphics[width=1in,height=1.25in,clip,keepaspectratio]{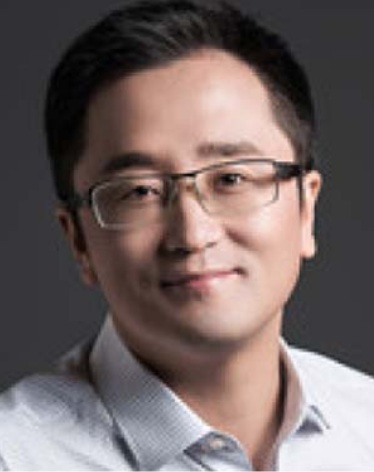}}]{Heng Tao Shen} (Fellow, IEEE) received the B.Sc. (with first class Hons.) and Ph.D. degrees from the Department of Computer Science, National University of Singapore, Singapore, in 2000 and 2004, respectively. He is currently with the University of Electronic Science and Technology of China, Chengdu, China. His research interests mainly include multimedia search, computer vision, artificial intelligence, and big data management. Prof. Shen is a member of Academia Europaea and a fellow of ACM, IEEE, and OSA. 
\end{IEEEbiography}

\end{document}